\def\paperID{0000}
\def\confName{CVPR}
\def\confYear{2024}

\def\paperTitle{M\&M3D: Multi-Dataset Training and Efficient Network for Multi-view 3D Object Detection}

\def\authorBlock{
    Hang Zhang \\
    Nanyang Technological University \\
    {\tt\small \{HZHANG067\}@e.ntu.edu.sg}
}

\newif\ifreview \newcommand{\review}{\reviewtrue}
\newif\ifarxiv 
\newif\ifcamera 
\newif\ifrebuttal 

\review 

\pdfoutput=1
\documentclass[10pt,twocolumn,letterpaper]{article}
\ifreview \usepackage{cvpr} \fi
\ifarxiv \usepackage[pagenumbers]{cvpr} \fi
\ifrebuttal \usepackage[rebuttal]{cvpr} \fi
\ifcamera \usepackage{cvpr} \fi

\usepackage{graphicx}	
\usepackage{amsmath}	
\usepackage{amssymb}	
\usepackage{booktabs}
\usepackage{times}
\usepackage{microtype}
\usepackage{epsfig}
\usepackage[table,xcdraw,dvipsnames]{xcolor}
\usepackage{caption}
\usepackage{float}
\usepackage{placeins}
\usepackage{color, colortbl}
\usepackage{stfloats}
\usepackage{enumitem}
\usepackage{tabularx}
\usepackage{xstring}
\usepackage{multirow}
\usepackage{xspace}
\usepackage{url}
\usepackage{subcaption}
\usepackage{xcolor}
\usepackage[hang,flushmargin]{footmisc}

\ifcamera \usepackage[accsupp]{axessibility} \fi

% ...

\ifarxiv  \fi

\newcommand{\R}[1]{{%
    \textbf{%
        \ifstrequal{#1}{1}{\textcolor{red}{R#1}}{%
        \ifstrequal{#1}{2}{\textcolor{blue}{R#1}}{%
        \ifstrequal{#1}{3}{\textcolor{magenta}{R#1}}{%
        \ifstrequal{#1}{4}{\textcolor{teal}{R#1}}{%
                           \textcolor{cyan}{R#1}%
        }}}}%
    }%
}}

\usepackage{xr-hyper}

\makeatletter
\newcommand*{\addFileDependency}[1]{
  \typeout{(#1)}
  \@addtofilelist{#1}
  \IfFileExists{#1}{}{\typeout{No file #1.}}
}

\makeatother

\definecolor{cvprblue}{rgb}{0.21,0.49,0.74}
\usepackage[pagebackref,breaklinks,colorlinks,citecolor=cvprblue]{hyperref}
\usepackage[capitalize]{cleveref}
\crefname{section}{Sec.}{Secs.}
\crefname{table}{Table}{Tables}
\crefname{figure}{Fig.}{Figs.}

\begin{document}
\title{\paperTitle}
\author{\authorBlock}

\twocolumn[{%
\maketitle
\begin{figure}[H]
    \hsize=\textwidth 
    \centering
    \includegraphics[width=\textwidth]{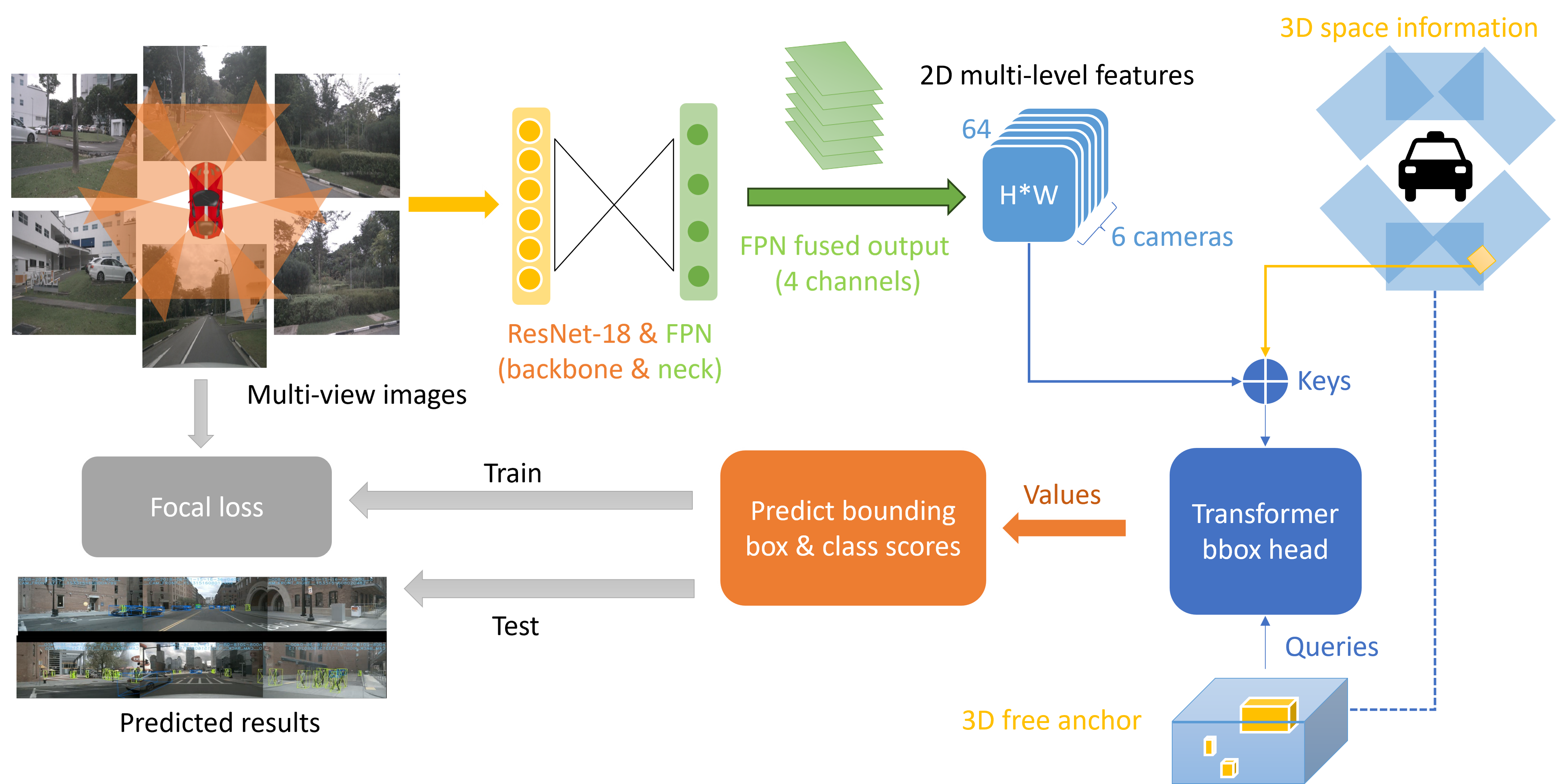}
    \caption{Overview of the new network structure. My designs focus on the 3D bbox prediction from customized parts, which generates 3D features by combining the 2D features and 3D position features through a position encoder. There are some customized network parts compared to the baseline model, which is 'Multi-level features in 2D', '3D space Information encoder', and 'Free anchor object query', feeding into the '3D bbox head based on a Transformer structure'. Then the orange part is customized for decoding the output to predicted bbox coordinates and class scores.}
    \label{model}
\end{figure}
}]

\begin{abstract}
In this research, I proposed a network structure for multi-view 3D object detection using camera-only data and a Bird’s-Eye-View (BEV) map. My work is based on a current key challenge - domain adaptation and visual data transfer. Although many excellent camera-only 3D object detection has been continuously proposed, many research work risk dramatic performance drop when the networks are trained on the source domain but tested on a different target domain. Then I found it is very surprising that predictions on bounding boxes (bbox) and classes are still replied to on 2D networks. Based on the domain gap assumption on various 3D datasets, I found they still shared a similar data extraction on the same BEV map size and camera data transfer. Therefore, to analyze the domain gap influence on the current method and to make good use of 3D space information among the dataset and the real world, I proposed a transfer learning method and Transformer construction to study the 3D object detection on NuScenes-mini and Lyft. Through multi-dataset training and a detection head from the Transformer, the network demonstrated good data migration performance and efficient detection performance by using 3D anchor query and 3D positional information. Relying on only a small amount of source data and the existing large model pre-training weights, the efficient network manages to achieve competitive results on the new target domain. Moreover, my study utilizes 3D information as available semantic information and 2D multi-view image features blending into the visual-language transfer design. In the final 3D anchor box prediction and object classification, my network achieved good results on standard metrics of 3D object detection, which differs from dataset-specific models on each training domain without any fine-tuning. The work is available at \href{https://github.com/Rango-Zhang-Hang/NTU-Visual-Recognition-via-Vision-Language-Model-Transfer.}{GitHub}.
\end{abstract}
\section{Introduction}
\label{sec:intro}
In recent years, the research and application of large-scale vision and language models have shown a rapidly growing trend, and some outstanding representatives include OpenAI's Contrastive Language-Image Pre-training (CLIP) \cite{radford2021learning} model and stable fusion model. This trend reflects the profound development in the fields of natural language processing (NLP) and computer vision (CV), and shows the ability of visual-language models to achieve major breakthroughs in 2D vision tasks, derived from combining image features and sequence context New concept of combination. The CLIP model, proposed by OpenAI in 2021, is a classic pioneer visual-language model. It has excellent multi-modal understanding capabilities and can process 2D images and text simultaneously to achieve cross-modal retrieval and classification tasks. The core idea is to align the embedding spaces of images and text with each other so that image and text descriptions can be compared in a common embedding space, thus achieving impressive cross-modal performance.

My work, M\&M3D, 'Multi-Dataset Training and Efficient Network for Multi-view 3D Object Detection', also borrows this idea to align the camera's multi-view images and the 3D space of the scene with each other, so that the network can use 2D image features and 3D spatial information to bring about a turnaround in the challenge of 3D object detection tasks. One of the key focuses of the project is to achieve efficient general feature extraction and application in the face of data constraints and limited computing resources, such as very limited labeled or unlabeled training data, to fully exploit the visual-language model structure potential. This requires me to design and develop new domain adaptation and 3D feature transfer technologies to fit the application of autonomous driving in actual scenarios, which means an unlimited target domain. Transferring 2D object detection to 3D or real scenes is currently a big challenge and research potential. This is a very promising practical project in the field of unsupervised or semi-supervised 3D visual recognition research.
\section{Related Work}
\label{sec:related}
\subsection{Bird’s-Eye-View}
Bird’s-Eye-View, or BEV in short, is a critical perspective in computer vision using a top-down view or overhead view, particularly in applications such as 3D object detection, and widely applied in autonomous driving. This section provides an in-depth exploration of the foundational research areas and influential works that have contributed to the advancement of BEV-based vision tasks. BEV-based methods tend to introduce the Z-axis error, resulting in poor performance for other 3D-aware tasks. This is a line of practical work of converting 2D image features into 3D predictions, using a 3D space transformer on BEV information. From CaDDN \cite{CaDDN}, BEVDet \cite{huang2021bevdet} and BEVDet4D \cite{huang2022bevdet4d}, it shows a solid proof for image-based 3D object detection explicitly predicts depth distribution to explore the new method in 3D space, from 2D attributes to real-world scenes. By performing successful visual attention, BEVFormer \cite{li2022bevformer} inspired a 2D-to-3D transformer with local attention in BEV grid and a parameter query. Taking ideas from 2D object detection, DETR3D \cite{wang2022detr3d} using 3D position presentations from a 2D work DETR \cite{2020detr}.

\subsection{Vision-based 3D object detection} 
Vision-based 3D object detection, generally for monocular or multi-view camera-only detection methods, has received increasing attention from researchers due to its rich semantics and low deployment cost and has seen rapid development in recent years. As another key perception task in autonomous driving, early approaches to 3D object detection are similar to 2D detection methods. These methods usually predict 3D bounding boxes based on 2D bounding boxes from camera images. The earl method started as RGB image detection from Mono3D \cite{mono3d} in 2016, using scene and other priors to collect semantic or shape proposals. Then \cite{roddick2018orthographic} shared the first idea of using BEV views and leveraging 2D detection work to minimize the 2D-3D visual gap. This work on 2D visual detectors as a starter and inspiration for following 3D visual recognition recently has become a trending approach to lower the training cost. Ssd-6d \cite{2017SSD-6D} and monocular-3D work like shape reconstructions \cite{2019Monocular3D}, adding geometric reasoning part\cite{2020MonocularDifferentiable} and SMOKE \cite{2020SMOKE} for one-stage network are continuing on monocular 3D object detection. Later, FCOS3D \cite{wang2021fcos3d} extends this paradigm to 3D
object detection and achieves great performance, inspiring many future multi-view 3D tasks.

Many researchers have worked on predicting objects directly from a single image view. However, limited data and a single viewpoint make it impossible to develop more complex tasks. Meanwhile, as some large benchmarks with more data and multiple viewpoints have been released together with multi-view datasets collected from autonomous driving industrial motivations, new perspectives are provided for the development of paradigms for multi-view 3D images. Thus, vision-based 3D object detection is gaining more attention \cite{ma2022visioncentric}. Due to rich semantic information and low cost for deployment in multi-view images, in the last few years, many efforts have been made to predict objects from multi-view image sets.
LSS \cite{2020lss} pioneered to proposal the multi-view features to BEV space. Based on these benchmarks, some multi-camera 3D object detection paradigms have been developed with good performance. A number of recent papers have already found that using Transformer and 2D target recognition frameworks can perform well on a single 3D dataset such as FCOS3D \cite{wang2021fcos3d}, BEVFormer\cite{li2022bevformer}, BEVFusion \cite{liu2023bevfusion}, DERT3D \cite{wang2022detr3d}, or supervised method such as BEVDet \cite{huang2021bevdet} and BEVDepth \cite{li2023bevdepth} to enhance the depth information. Then 3D convolutions and domain-specific heads are used to detect objects in both indoor and outdoor scenes. to transform 2D multi-view features into BEV representation. Transformer-based methods can gain benefits from 2D well-developing modeling tricks with additional training augmentations in 3D information. DETR3D \cite{wang2022detr3d} and PETR \cite{liu2022petr} propose 3D position-aware encoding, which greatly improves the performance of DET3D.

\section{Method}
\label{sec:method}

\subsection{3D Multi-Datasets Training}
In this section, we will briefly introduce and analyze two important types of 3D datasets: NuScenes \cite{caesar2020nuscenes} and Lyft \cite{lyft}. Our main research goal is to analyze the domain gap between them. Both datasets are designed to represent real-world road scenes and provide rich sensor data. Here we only take camera data to the table, the multi-view data, to simulate the automatic driving scenes. However, although they are all oriented towards similar 3D visual tasks, there is a domain gap between them, which has an important impact on the generalization ability and performance of the networks. T

\begin{figure}[t]
        \begin{subfigure}{0.45\linewidth}
            \includegraphics[scale=0.2]{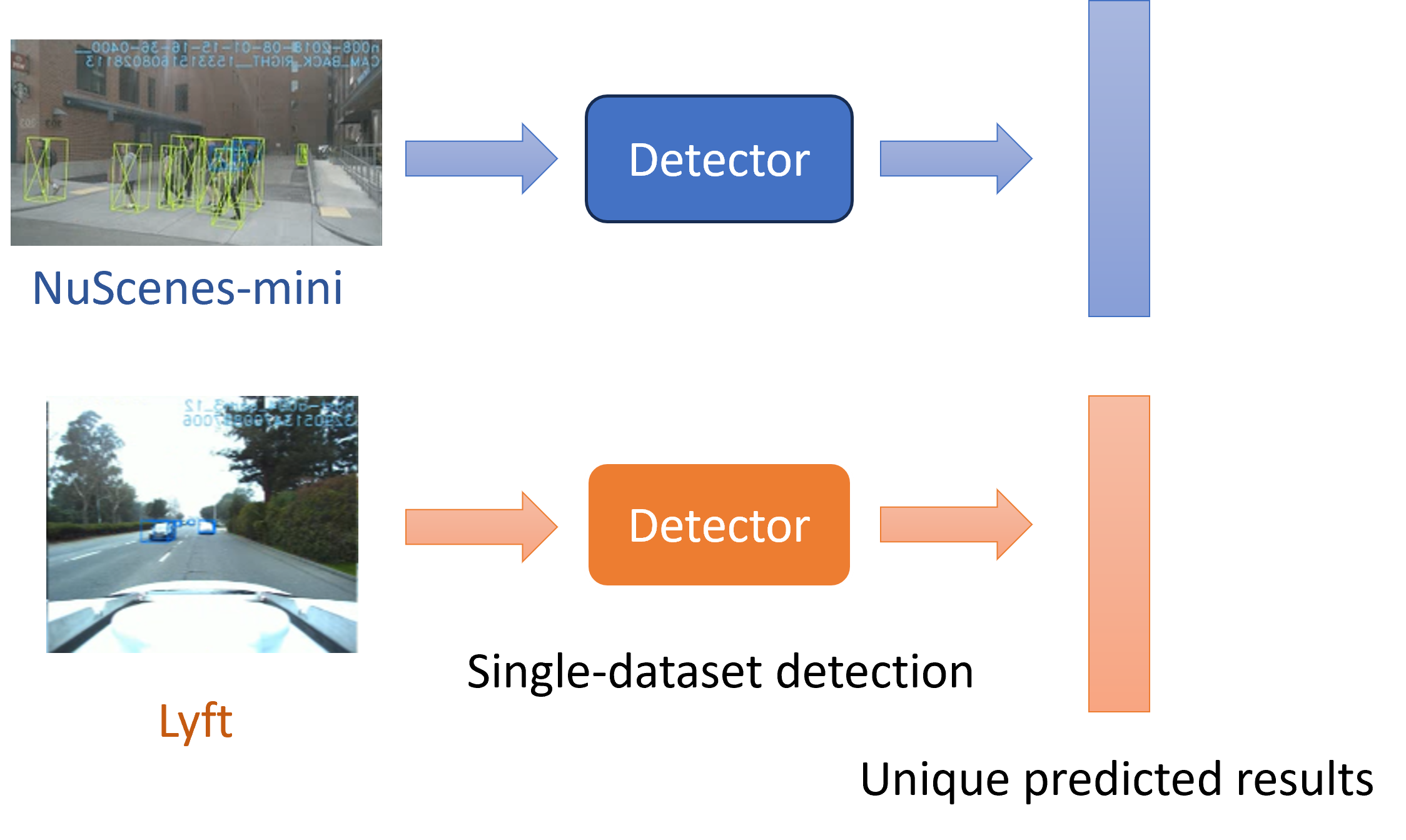}     
            \caption{Normal single dataset training}
        \end{subfigure}
        \hfill
        \begin{subfigure}{0.45\linewidth}
            \includegraphics[scale=0.15]{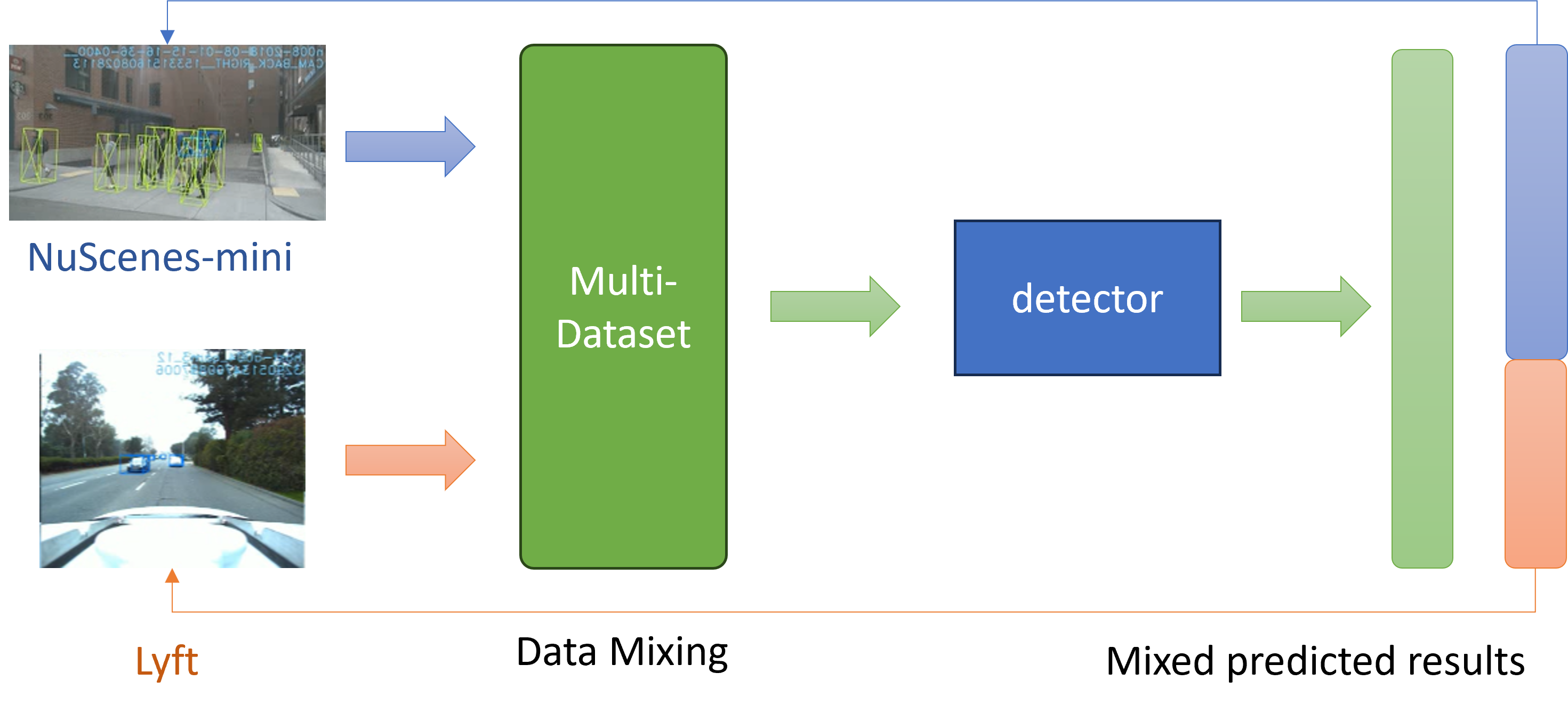}     
            \caption{Multi-dataset training with data mixing}
        \end{subfigure}
    \caption{Detectors represent the network part trained on source data. And detectors output their predictions or classifications. In the first one $(a)$, it is a standard detector trained and validated on one dataset. The multi-dataset training $(b)$ here is a mixed-dataset associated with datasets from different domains, before the 3D object detection network training.}
\end{figure}

\subsubsection{Domain Gap between NuScenes-mini and Lyft}
The domain gap that exists between different data sets is an indisputable fact. Different cameras can produce different domains in a 3D dataset, mainly because of differences in the way they capture and represent visual information. Different cameras use different sensors that have unique characteristics and technologies, including sensitivity to light, noise levels, and color representation, which may lead to differences in the visual data they capture.
A camera also has a variety of shooting settings, and different settings can cause that camera to capture more and different details, such as variations in lighting, resolution, and other specialized parameters. The level of detail affects the way objects and scenes are presented in the image. Even more, the camera may have different color profiles or color calibrations during data production, which can affect the way colors are rendered in the image. This may result in variations in color accuracy and saturation.

\subsubsection{Multi-dataset Training in 3D Object Detection}
This approach has been successful in 2D visual recognition. Then we expect to borrow this form and hope that it is also a practical approach in scenarios where 3D object detection tasks are challenging. As we mentioned above and in the related work section, some of the best 3D models or network designs, such as BEVFormer \cite{li2022bevformer}, have shown very strong performance on the NuScenes dataset alone. But it is very challenging to improve upon the existing results with the limited computational resources I have and it is very challenging to improve on the existing results. That's why we mention the use of 'Multi-dataset Training' to improve the network's performance because I want the network to be able to efficiently learn more information about the 3D image from a new perspective while leveraging the properties of these excellent models, and it can transfer to a new target domain. It is important to note that this approach is currently a very promising direction in 3D, and any theoretically feasible way of pre-data is architecture-independent.

\begin{figure}[h]
\centering
\includegraphics[width=0.9\linewidth]{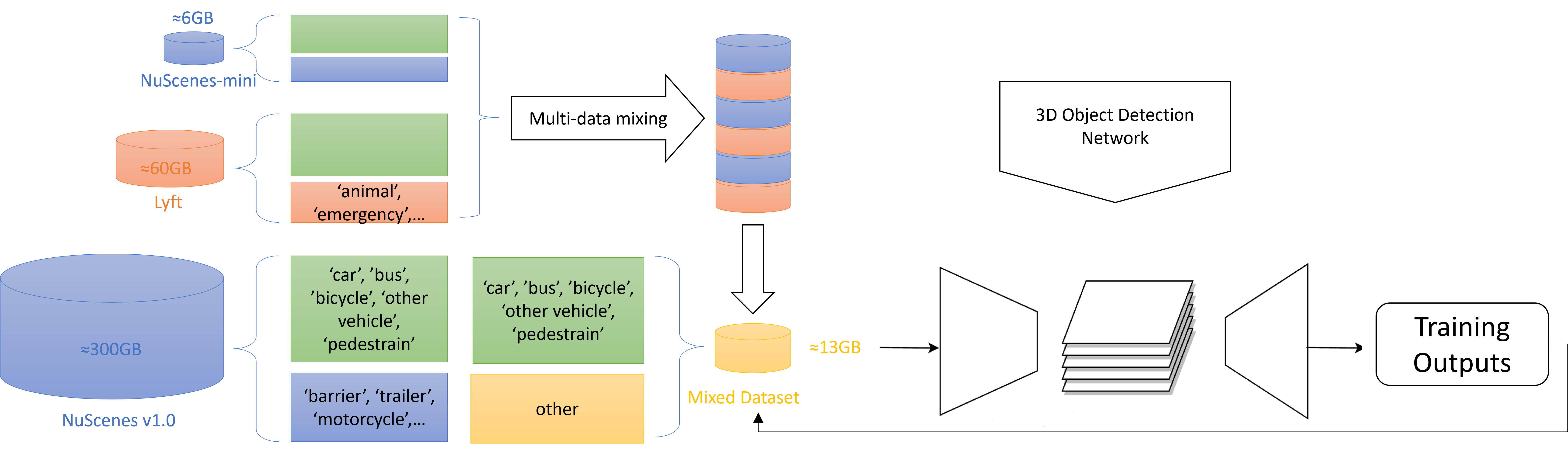}
\caption{Multi-Dataset Training for 3D datasets. We prepossessed our multiple 3D datasets to a joint dataset with a mixed label set containing the shared classes and others. At the start of a training epoch, the samples are scanned from each dataset equally. Then the 3D object detection network uses the scanned samples as input. This figure represents the training procedure when NuScenes-mini and Lyft are the source datasets. NuScene v1.0 is also shown as a comparison in later sections.}
\label{fig3.2}
\end{figure}

\textbf{Mixing Data from Different Dataset} is a key part of our domain adaptation method. In multi-dataset training, the model can train multiple datasets simultaneously, some of which may be relevant to the target task. In this case, the model can improve its performance on the target task by learning on multiple data sets, thereby achieving the effect of transfer learning. Instead of using a single 3D dataset, I aim to take advantage of the adaptation to the unseen scenes or a new type of camera. Here for my NuScenes-mini and Lyft, and also for possible future research, I named the multi-dataset $D={d_{i}, i=0,1...n}$, where $d_i$ represents one of the pure 3D datasets I used to mix in with various data distriution. With $d_i = {(B_{i}, CLS_{i}), POS_i}$, the key data distributions of each class and objects are represented by annotations with a bounding box set $B_i$ with $bbox$ with 3D coordinates, and $CLS_i$ with related labels $cls$. Also, $POS_i$ is used to represent other fixed 3D information such as BEV map size, and camera instincts information that every dataset uses for data format transfer. It is very obvious that NuScenes-mini and Lyft hold different dataset sizes and formats, this imbalance number of the camera images in each dataset needs an aligned simple mixing method. Therefore, as shown in Fig \ref{fig3.2},  I shuffled and regrouped the images, keeping the 3D position relationship, and a mixed dataset with six positions images set with the diversity of NuScenes-mini and Lyft, will be used for further training and experiments. Various scenes will be included in each epoch that the network would be forced out of overfit and could be used in the data transfer learning.  

\textbf{Labels for Mixed Dataset} is a practical challenge I met trying to solve the domain gap of datasets. Each self-driving dataset uses its own set of labels, corresponding to objects with different semantic hierarchies. For example, the NuScenes dataset combines a collection of object classes, including human, vehicle, etc., with potential subclasses represented in the form human.pedestrain.adult. Lyft's open dataset annotations will contain simple object classifications, but not more refined classifications. After a brief consideration, the label differences shown in Fig \ref{} only contain object labels, while the hidden subclasses continue to follow the structure of the source dataset. As analyzed before, the unreconciled parts of several datasets will more or less destabilize the training because the datasets do not have the same class nomenclature. Additionally, I would like to have a more unified set of statistical methods. 

\subsection{M\&M-3D}
\begin{figure}[h]
\centering
\includegraphics[width=\linewidth]{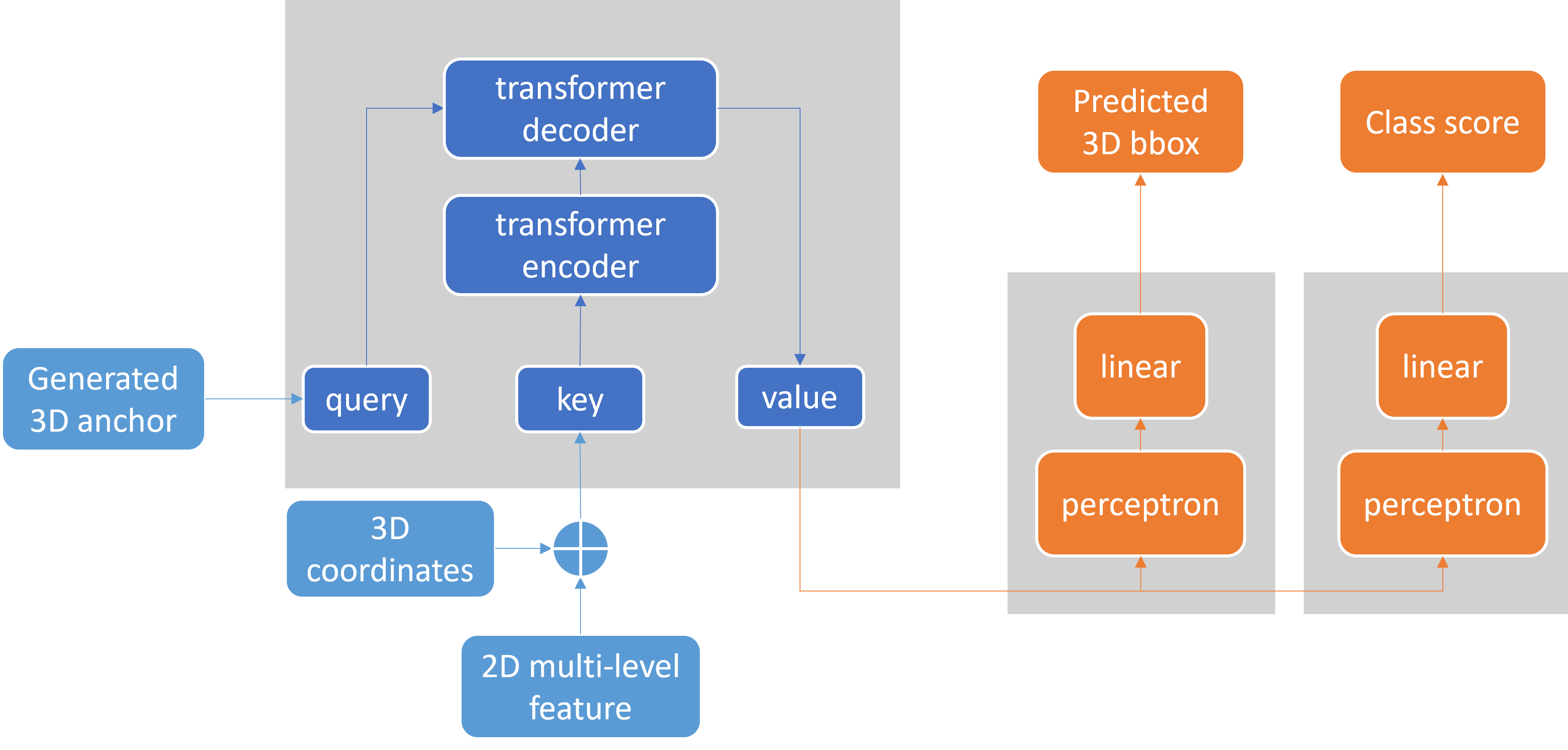}
\caption{Transformer network design. Taking the processed 2D multi-level features with extra position information on 3D coordinates, they are added together to the transformer encoder as keys. The query is the anchor generated by an MMDet3D anchor head updated to the transformer decoder. The output value from the decoder is then fed to two separate networks for predicted bbox and class.}
\label{transformer}
\end{figure}

The entire transformer network involved in Fig \ref{trans} can be taken a brief look at. Although transformers have been widely used in the language and vision domains in recent years because of their powerful visual-language transfer, their actual deployment for 3D object detection still requires additional modifications. In addition, the 2D image features output from the FPN also needs to be further modified to 2D multi-level features. 

\begin{figure}[htbp]
	\centering
	\begin{minipage}{0.49\linewidth}
		\centering
		\includegraphics[width=0.9\linewidth]{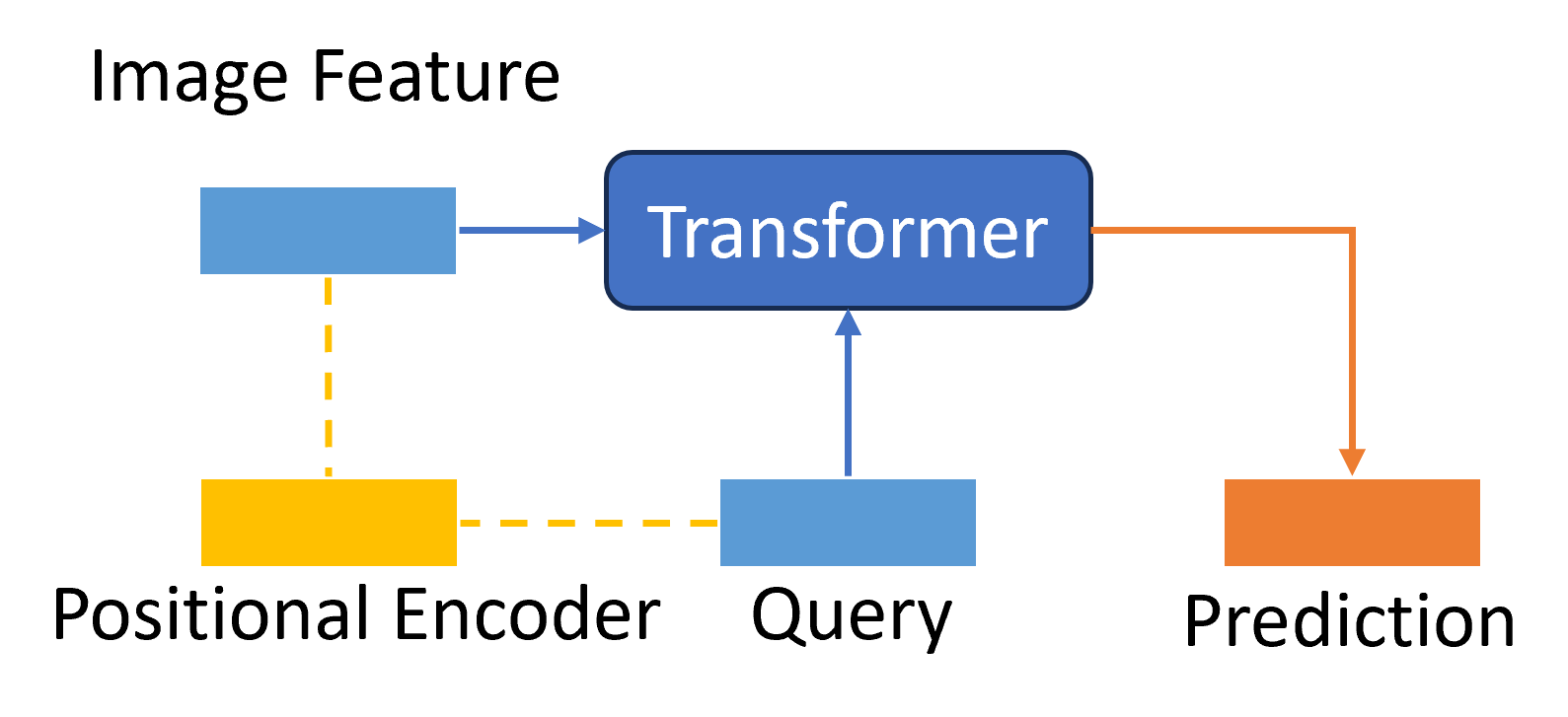}
		\caption{A general attention structure}
		\label{1}
	\end{minipage}
	\begin{minipage}{0.49\linewidth}
		\centering
		\includegraphics[width=0.9\linewidth]{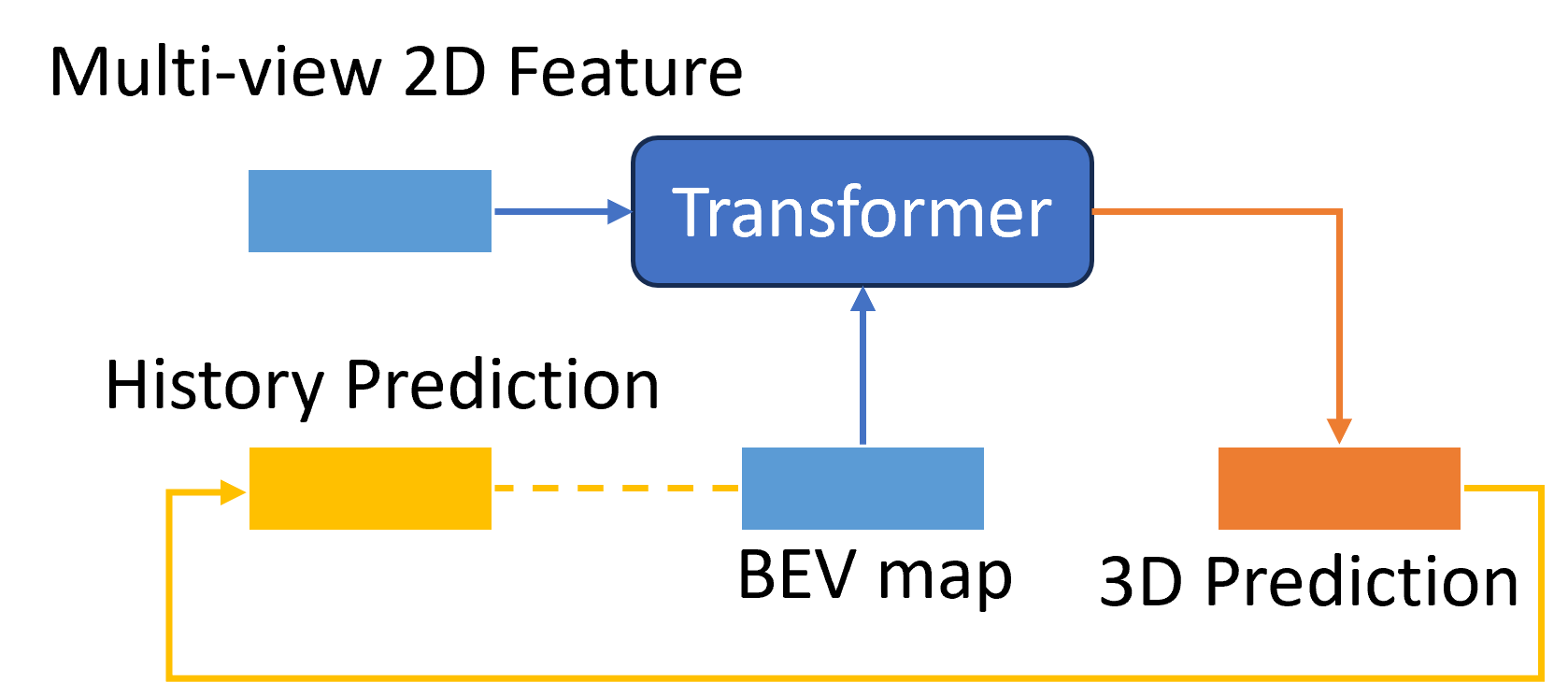}
		\caption{BEVFormer}
		\label{2}
	\end{minipage}
	
	\begin{minipage}{0.49\linewidth}
		\centering
		\includegraphics[width=0.9\linewidth]{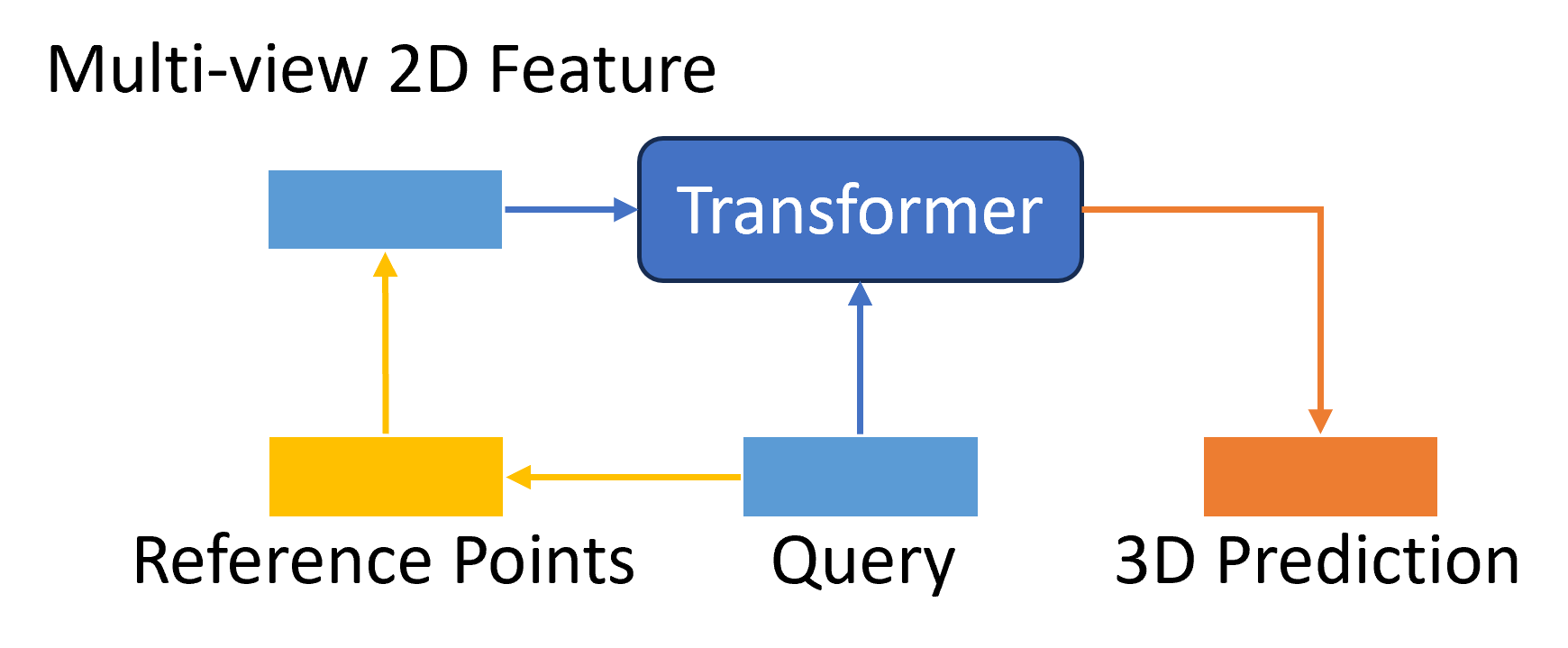}
		\caption{DETR3D}
		\label{3}
	\end{minipage}
	\begin{minipage}{0.49\linewidth}
		\centering
		\includegraphics[width=0.9\linewidth]{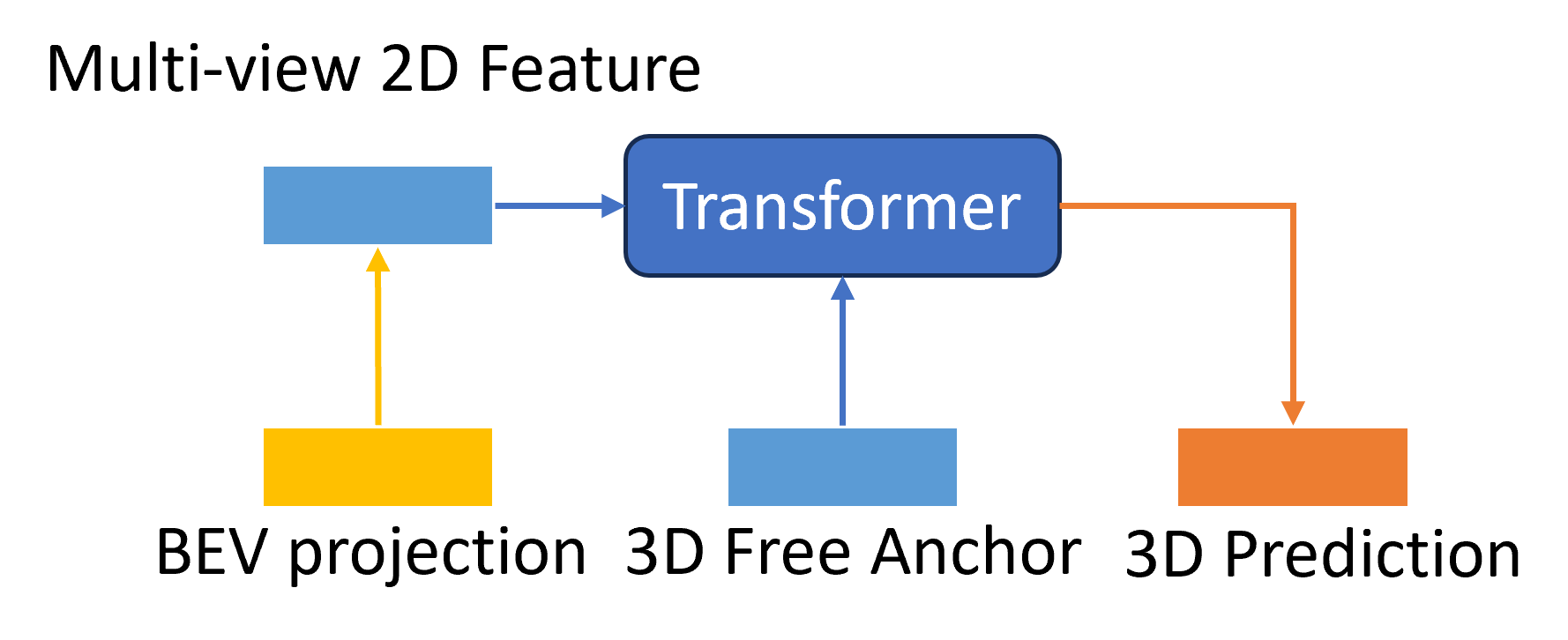}
		\caption{New Design}
		\label{4}
	\end{minipage}
\caption{Attention in 3D object detection work.}
\label{4mini}
\end{figure}

I don't mean the 3D attention is brought out here as the first time for 3D object detection. However, as shown in Fig.\ref{4mini} existing 2D-to-3D transformer works remain many future challenges. The query was represented by BEV grid-size parameters (BEVFormer) or reference points in camera images (DETR3D). The initiated inaccurate coordinates may lead the network to perform weirdly from the training features. Also, BEVFormer used the space-cross part of the BEV map but didn't connect the global location; DETR3D introduced the related points but repeatedly projected them, which is less effective if the attention is enough. Combining their advantages is a tryout and the main idea of this new design. So in my new design, I use related 3D coordinates among the 3D space to imply the possible real-world objects and also give a global space region to improve the positional information quality in 2D images.

\subsubsection{Multi-level feature process} \label{subsec-mlvl}

\begin{figure}[h]
\centering
\includegraphics[width=\linewidth]{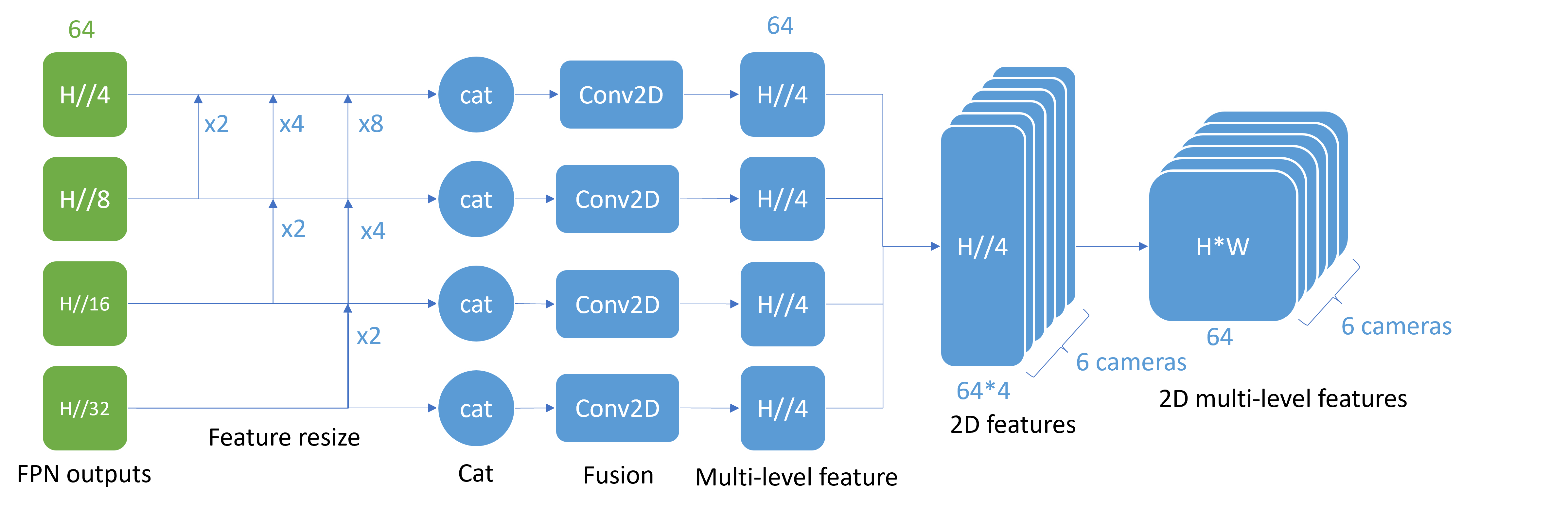}
\caption{Neck fusion: from FPN outputs to multi-level 2D features. }
\label{mlvl feat}
\end{figure}

The FPN outputs are 4 channel with size $6*64*h_{i}*w_{i}$ for the $i^{th}$ channel, where $h_{i}, w_{i} = \frac{H}{2^{i+1}}, \frac{W}{2^{i+1}} i \in (1,2,3,4)$. Then for each channel, the network upscales every other channel with smaller feature sizes and uses a cat function to add them together. Next, as four features with same feature size $\frac{H}{4}, \frac{W}{4}$ but different channels, I introduced four fusion Conv2D with different in channels but same 64 out channels. Until now, the network converts the FPN outputs of different sizes of features to a multi-level feature with size $4*64*\frac{H}{4}* \frac{W}{4}$. To fit the following Transformer and the related 3D information, a final reshape is needed to output the 2D multi-level feature in size $6*64*H*W$. Noted that there are 6 camera images fed to the network at the same batch, forming the 3D scenes.

\subsubsection{3D Anchor as Customized Query} \label{subsec-query}
As mentioned before, the Transformer structure will be used for the bbox prediction work. As the integral query in my Transformer network, I chose to customize the 3D anchor box as the query for the task of 3D object detection. The functions and the class I used here to generate the promoted anchors are from the class 'Anchor3DRangeGenerator' \cite{zhang2019freeanchor} provided by MMDet3D \cite{mmengine2022}. 

\begin{figure}[h]
\centering
\includegraphics[width=0.9\linewidth]{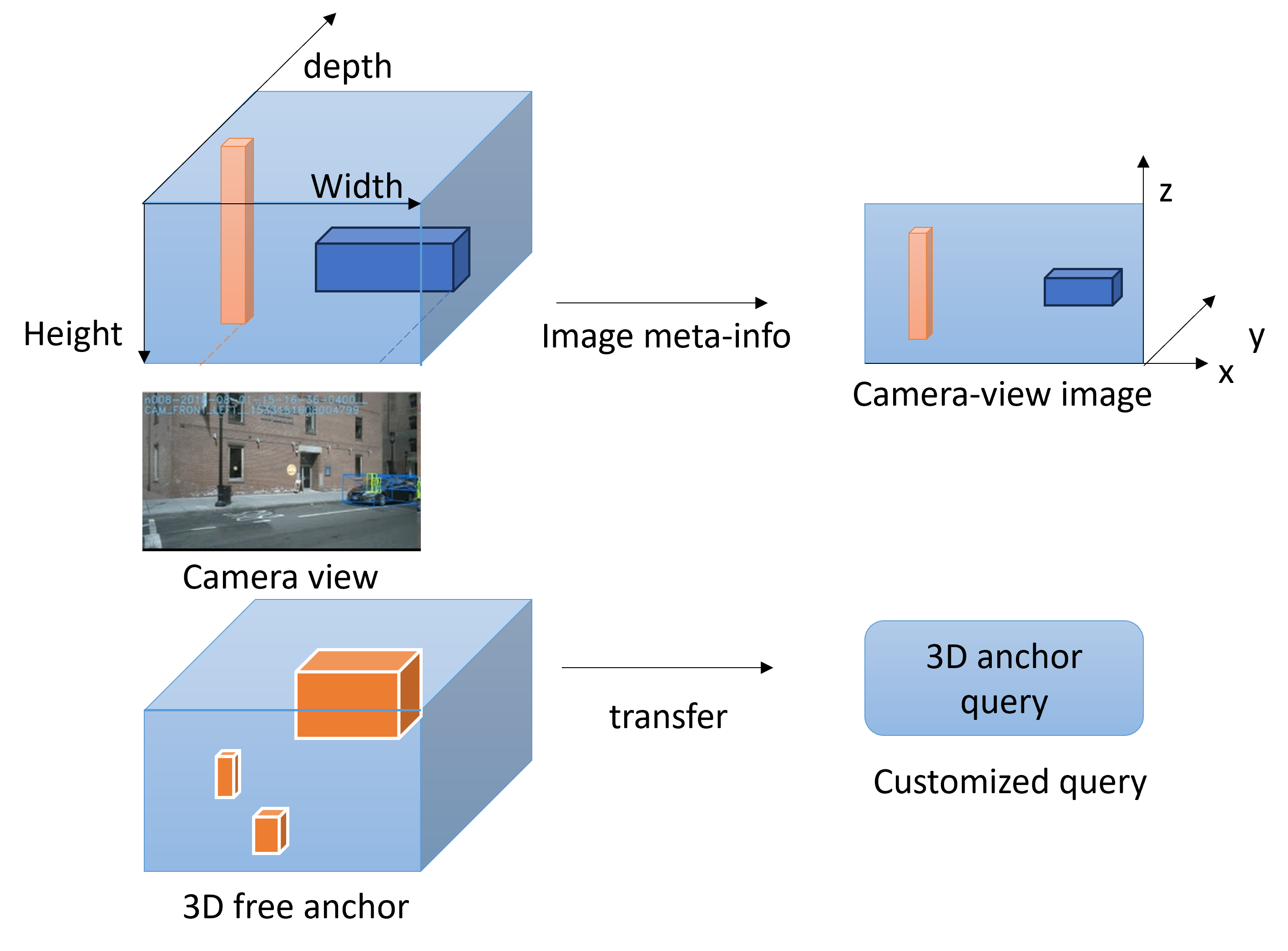}
\caption{From 3D anchor to Transformer query.}
\label{query}
\end{figure}

As shown in Fig \ref{query}, it generates the center by evenly distributing the predefined size of anchor frames in the minimum and maximum range according to the size of the image. If we give the pre-provided 3D anchor type $B \in (cx, cy, cz, w, h, l, k)$ in camera format, then the function I named $F$ will give information about all possible 3D anchors in a known camera-aware field of view $P \in (X_{min}, X_{max}, Y_{min}, Y_{max}, Z_{min}, Z_{max})$, which is subsequently passed through the meta info of the datasets (also the camera intrinsic and extrinsic information to know the camera view range $W, H, D$ and twisted angle $K$) into our common visual bbox $(cx, cy, cz, lx, ly, lz)$. 

\subsubsection{3D Coordinates as Positional Encoder} \label{subsec-pos}
Two points worth noting, however, are that first, we can only obtain 3D coordinates from 2D images. Secondly, the 3D position will only be added to the 2D multi-level feature (Transformer input), not to the anchor as Query, because the 3D anchor already has a 3D position, so there is no need to add the same content again. This means that we need to do some processing on the initial $P_{2D}$ to get the final $P_{3D}$, and my expression for the 3D position needs to end up in the same format as the 2D multi-level feature, but with the same content as the anchor.

\begin{figure}[h]
\centering
\includegraphics[width=\linewidth]{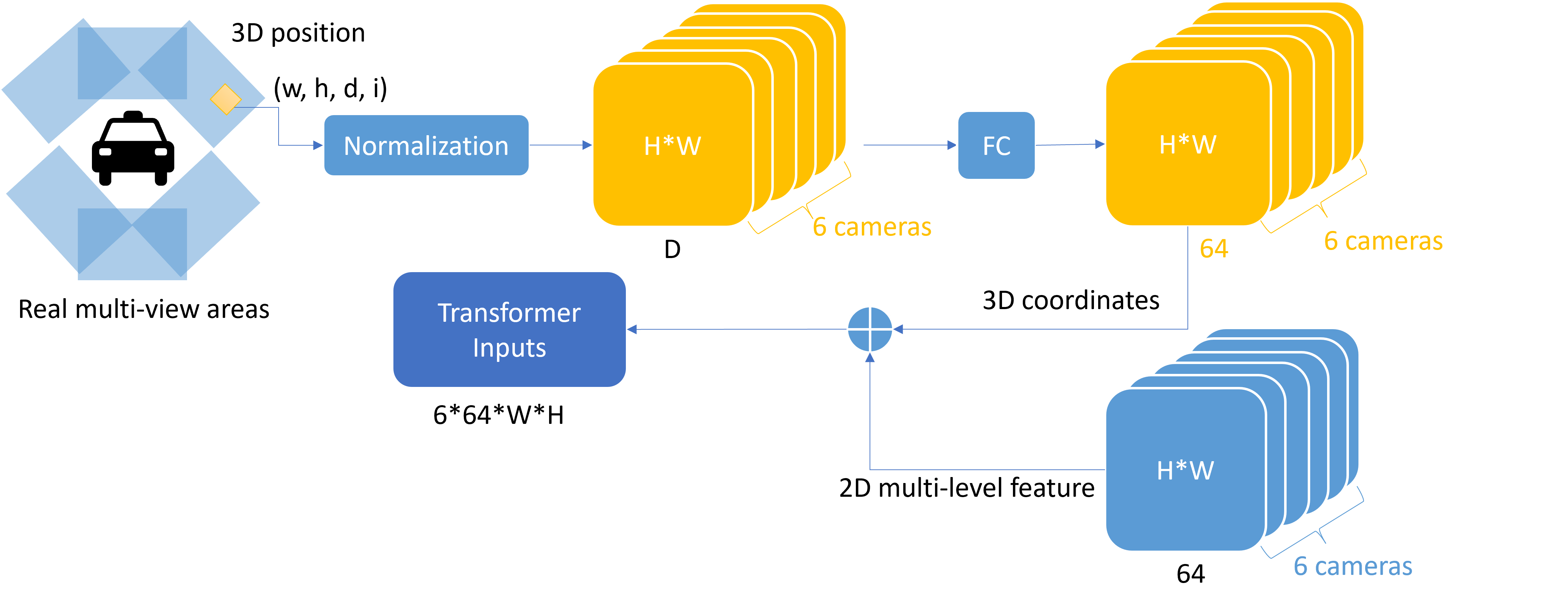}
\caption{3D coordinate information from related 3D position in multi-view images.}
\label{3Dpos}
\end{figure}

As shown in Fig \ref{3Dpos}, I decided to start with the pixel points of the 2D image, as the picture pixel points directly represent its position in our 'multi-view images'. Firstly, a rough representation of the position of pixel point $i$ in a set of 6 camera images can be written as $(x_{i}, y_{i}, j)$, where $j$ means the $j$th of the 6 views from 'FRONT LEFT', 'FRONT', 'FRONT RIGHT', 'BACK LEFT', 'BACK', 'BACK RIGHT'. Obviously, this representation is not feasible, because there is no way to represent the 3D information in $x y$ here, and it is not consistent with the previous anchor. Then, referring to the previous anchor frame generation, we express any point in a set of 6 images like this: $(w_{i}, h_{i}, d_{i}, j)$, where $w_{i} \in (W_{min}, W_{max}) h_{i} \in (H_{min}, H_{max}) d_{i} \in (D_{min}, D_{max})$. Then I further normalize the pixel coordinates in order to make sure that such 3D information also works for different $W, H, D$ and to reduce the amount of computation. Then a pixel point in the $j^{th}$ view can be represented as

$$(\frac{w_{i}-W_{min}}{W_{max} - W_{min}}, \frac{h_{i}-H_{min}}{H_{max} - H_{min}}, \frac{d_{i}-D_{min}}{D_{max} - D_{min}})$$

, where $w_{i} \in (W_{min}, W_{max}) h_{i} \in (H_{min}, H_{max}) d_{i} \in (D_{min}, D_{max})$

So now, we get 3D coordinates of size $6*D*H*W$. In order to be consistent with the 2D multi-level feature format, we add a fully connected layer to make the final 3D coordinates $6*64*H*W$.

\subsubsection{Transformer Bounding Box Head}
Although the performance of conv2d has been confirmed by a lot of excellent work, it has spatial invariance when processing features at different locations. This means that Conv2D network cannot directly consider the 3D information of a specific location, which has a very large negative impact on our 3D target detection and 3Dbbox prediction. In addition, Conv2D cannot explicitly express the conversion from image features to bbox.

\begin{figure}[h]
\centering
\includegraphics[width=0.9\linewidth]{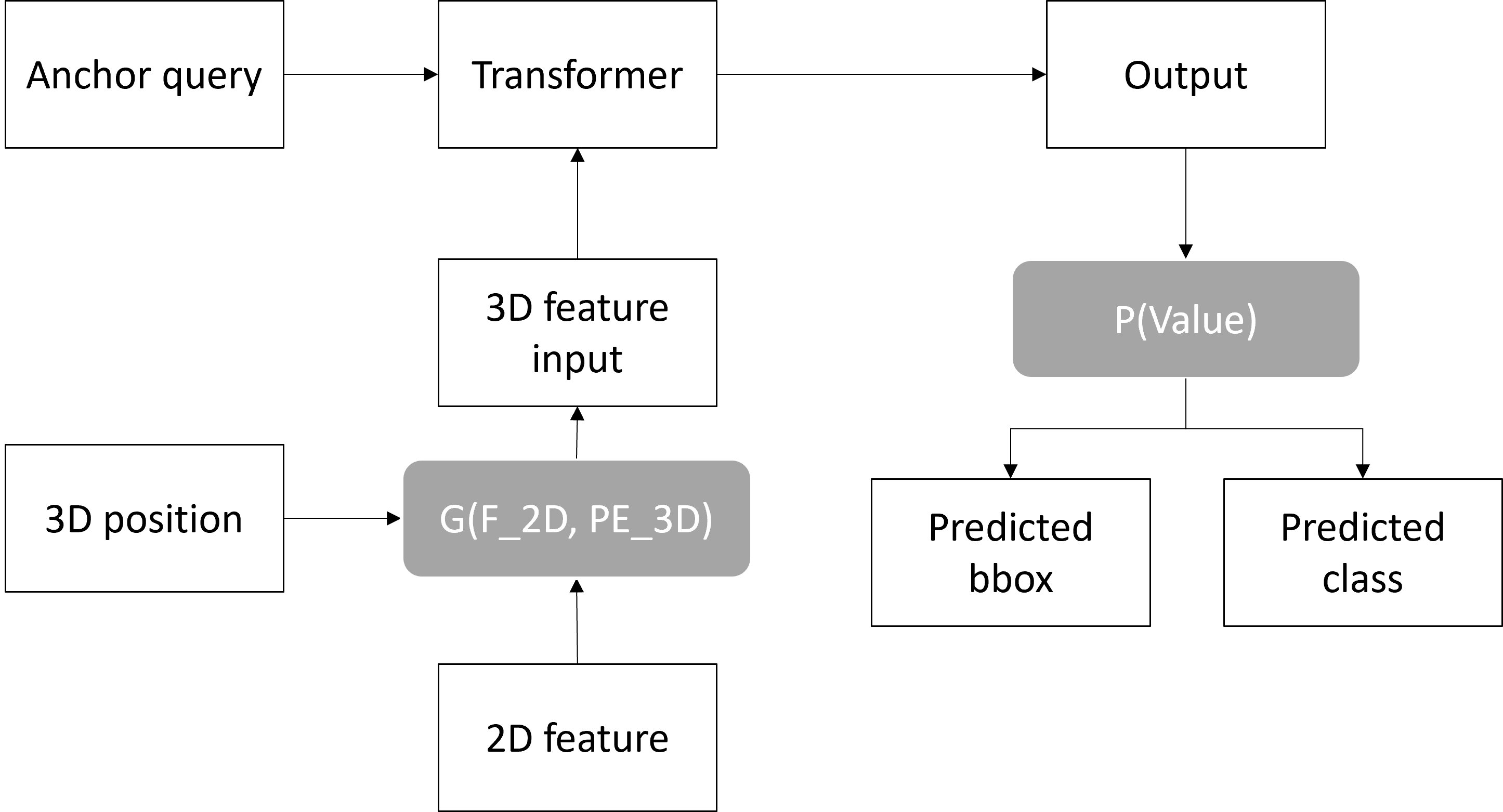}
\caption{Simple transformer network with inputs and outputs. The networks in grey represent the functions I used that one combines the 2D multi-level features with 3D coordinates, and another predicts the bbox and class from the Transformer output}
\label{trans}
\end{figure}

Now, let me explain how some of the data I customized earlier will enter this Transformer network.

\textbf{Transformer} network contains the input and output that will processed by it including 2D image multi-level features $f_i \in F_{2D}$, 3D coordinates $pos_i \in PE_{3D}$, anchor box query $q_i \in Q_ {Anchor}$, $v_i \in Value$, and the final result $(bbox_i, cls_i) \in (Pre\_bbox, Class) = p(Value)$. The self-attention mechanism in Transformer allows the network to automatically learn the relationship between features and notice features at different locations. Specifically, the network combines the input with sequence $f_i^{3D} \in F_{3D}=g(F_{2D}, PE_{3D})$ with the query to obtain each feature and each anchor attentional correlation between boxes. In other words, the network will automatically learn the relationship between features, notice features at different locations, and calculate the attention score $Attention(F_{3D}, Q_{Anchor}, Value)$ between them. When predicting the task, we combine the attention score $Attention(F_{3D}, Q_{Anchor}, Value)$ and the input $f_j$ to get the final prediction result $(bbox_j, cls_j)$. The grey networks in Fig \ref{trans} are introduced separately in the input and output parts below.

\textbf{Inputs} are 3D features combining the 2D multi-level features and 3D coordinates. The 3D features $f_{i}^{3D} \in F_{3D}$ could be extend as  $f_{i}^{3D} \in R^{6*H*W}$ where $i=0,1...63$. Also, the 2D multi-level features are $f_{i} \in F_{2D}=R^{6*H*W}$ where $i=0,1...63$ and 3D coordinates $pos_i \in PE_{3D}=R^{6*H*W}$ where $i=0,1...63$. However, as in position part \ref{subsec-pos}, the original $p_i \in PE_{3D}=R^{6*H*W} where i=0,1...D-1$ where $D$ is the depth in known 3D information, which transposed into a FC layer then we have $pos_i$. For the combination of features and positional encoder, I cat them of the same size together to be the 3D features.

\textbf{Outputs} from the network will be sent to another network to predict bbox and class results. In the structural sense of Transformer, Output is actually Query at the next moment. But in my simple Transformer, the output is direct, containing the corresponding 3D features. But for a single predicted data $j$, after the Transformer outputs the output, further processing is required to obtain the predicted bbox and class information from the same set of outputs. As shown in Fig \ref{transformer}, here I have designed two identical ‘Perception + Linear’. Because here the 3D output is converted to 3D bbox and category, so only simple regression and classification functions are needed.

\section{Results}
\label{sec:results}

\subsection{Experiments Setup} \label{Experiments Setup}
\textbf{Implementation Details} followed the standard as others methods and work. The network structure follows the BEVFromer settings. The initial image size is the smallest $256x704$ considering the network's efficiency, with data augmentations including random scaling, random flipping, and random rotation. For the 2D image feature extraction, I used ResNet-18 with PyTorch pre-trained checkpoint, FPN for multi-scale features with sizes of 1/4, 1/8, 1/16, 1/32, and the channel dimension of 64. For the 3D information on datasets, the BEV map size is the default $200x200$, where the 3D space ranges from $-51.2m$ to $51.2m$ for $X$ and $Y$ and $-5m$ to $5m$ for $Z$. For the training, I used one 12GB GTX2080Ti with 100GB disk space to train the model. The configuration contains 10 epochs(2x schedule in MMDet3D) training in 16 hours with a base learning rate $2x10^{-4}$ and the AdamW optimizer with a weight decay of 0.01. No data
augmentation methods are used for the test.

\textbf{Dataset \& Evaluation Metrics} for the 3D detection task are same with others. I used nuScenes Detection Score (NDS), mean Average Precision(mAP), mean Average Translation Error (mATE), mean Average Scale Error (mASE), mean Average Orientation Error(mAOE), mean Average Velocity Error (mAVE), mean Average Attribute Error (mAAE) as the result scores shown in result tables for NuScenes-mini, Lyft and the mixed dataset.

\subsection{Results}
\subsubsection{Benchmark Results}
This is a part summarizing other pioneer and excellent work that also uses camera data. Comparing their adjustments on the backbone, and image size, their results show more information beyond my limited research.
\begin{table}[h]
\centering
\resizebox{\columnwidth}{!}{
\begin{tabular}{ccccccc}
\\ \hline
\multirow{2}{*}{Method} & Modality    & \multirow{2}{*}{Backbone} & \multirow{2}{*}{Image Size} & \multirow{2}{*}{Epochs} & \multicolumn{2}{c}{3D Object Detection} \\
                        & Camera Only &                           &                             &                         & NDS ($\uparrow$)           & mAP ($\uparrow$)           \\ \hline
FCOS3D \cite{wang2021fcos3d}                  & yes         & ResNet-101                & 1600x900                    & 48                      & 0.372                    & 0.295                    \\
BEVFormer-S             & no          & ResNet-50                 & -                           & 48                      & 0.448                    & 0.375                    \\
BEVFormer \cite{li2022bevformer}               & no          & ResNet-101                & 1600x900                    & 48                      & 0.517                    & 0.416                    \\
BEVDet \cite{huang2021bevdet}                 & yes         & ResNet-50                 & 704x256                     & 48                      & 0.372                    & 0.286                    \\
BEVDet                  & yes         & ResNet-50                 & 704x256                     & 48                      & 0.379                    & 0.298                    \\
BEVDet                  & yes         & ResNet-101                & 704x256                     & 48                      & 0.381                    & 0.302                    \\
BEVDet                  & yes         & Swin-Tiny                 & 704x256                     & 48                      & 0.392                    & 0.312                    \\
BEVDet                  & yes         & ResNet-50                 & 1056x384                    & 48                      & 0.389                    & 0.318                    \\
BEVDet                  & yes         & ResNet-101                & 1056x384                    & 48                      & 0.396                    & 0.330                    \\
BEVDet                  & yes         & Swin-Tiny                 & 1056x384                    & 48                      & 0.410                    & 0.333                    \\
BEVDet                  & yes         & Swin-Tiny                 & 1408x512                    & 48                      & 0.417                    & 0.349                    \\
DETR3D \cite{wang2022detr3d}                 & yes         & Swin-Small                & 1600x900                    & 48                      & 0.374                    & 0.303                    \\
DETR3D-CBGS             & yes         & Swin-Small                & 1600x900                    & 48                      & 0.434                    & 0.349                    \\
PETR \cite{liu2022petr}                   & yes         & ResNet-50-DCN             & 1056x384                    & 48                      & 0.381                    & 0.313                    \\
PETR                    & yes         & ResNet-101-DCN            & 1408X512                    & 48                      & 0.421                    & 0.357                    \\
PETR                    & yes         & Swin-Tiny                 & 1408X512                    & 48                      & 0.431                    & 0.361                    \\
BEVDepth \cite{li2023bevdepth}               & no          & ResNet-50                 & 704x256                     & 48                      & 0.475                    & 0.351                    \\
BEVDepth                & no          & ResNet-101                & 1408X512                    & 48                      & 0.535                    & 0.421                    \\
\textbf{BEVDepth}                & no          & ResNet-101-DCN            & \textbf{1408X512}                  & 48                      & \textbf{0.538}                    & \textbf{0.418}                    \\ \hline
Network               & Camera Only          & Backbone            & Image Size                    & 10                      &\multicolumn{2}{c}{NuScenes-mini} \\ \hline
My Design               & yes          & ResNet-18            & 704x256                    & 10                      & 0.4637                    & 0.3767 \\ \hline
\end{tabular}}

\caption{Other great work before May 2023 for 3D object detection using data from camera sensors. The NuScenes val results are directly from their papers. \textbf{My Design} represents the multi-dataset training + Transformer on NuScene-mini val without fine-tuned. However, NuScenes-mini val set is way to small compare to NuScenes val set, the results here are only for the demonstration.}
\label{baseline}
\end{table}

Through comparison, it is found that existing work confirms that even on a large-scale data set such as NuScenes v1.0, just replacing a larger backbone and inputting a larger image size has a significant impact on the final 3D object detection score. My work is inspired by it. First, these works proved the feasibility of transferring 2D visual recognition to 3D visual recognition, and then some work confirmed the importance of 3D spatial features such as BEV for 3D object recognition, especially models trained using only multi-view camera data. This is also the main reason why my network can process efficiently using only multi-view camera 2D images and BEV maps, combined with the modified Transformer structure.

\subsubsection{My Results on NuScenes-mini and Lyft}
Table \ref{nusc-new} and Table \ref{lyft-new} show my network prediction results with multi-dataset training. Although Lyft is not a popular research dataset, I still want to use the predictions on Lyft to show the impacts form my new network design. 
\begin{table}[h]
\centering
\resizebox{\columnwidth}{!}{
\begin{tabular}{ccccccc}
\hline
\multicolumn{7}{c}{NuSenes-mini Results}                                   \\ \hline
Object Class         & AP ($\uparrow$)& ATE ($\downarrow$)& ASE ($\downarrow$)& AOE ($\downarrow$)& AVE ($\downarrow$)& AAE ($\downarrow$)\\ \hline
Car                  & 0.609  & 0.467  & 0.136  & 0.205  & 0.692  & 0.240  \\
Bicycle              & 0.338  & 0.584  & 0.228  & 0.413  & 0.345  & 0.053  \\
Motorcycle           & 0.384  & 0.635  & 0.188  & 1.014  & 0.957  & 0.214  \\
Truck                & 0.399  & 0.553  & 0.198  & 0.242  & 0.662  & 0.238  \\
Bus                  & 0.431  & 0.611  & 0.176  & 0.221  & 0.945  & 0.283  \\
Pedestrian           & 0.441  & 0.672  & 0.232  & 0.944  & 0.665  & 0.293  \\ \hline
\multicolumn{1}{l}{mAP for Common 6} & \multicolumn{6}{c}{0.4370}  \\ \hline
Construction Vehicle & 0.147  & 0.962  & 0.408  & 0.896  & 0.122  & 0.383  \\
Trailer              & 0.153  & 0.903  & 0.214  & 0.571  & 0.469  & 0.131  \\
Traffic Cone         & 0.477  & 0.492  & 0.344  & nan    & nan    & nan    \\
Barrier              & 0.436  & 0.712  & 0.241  & 0.119  & nan    & nan    \\ \hline
NDS ($\uparrow$) & mAP ($\uparrow$) & mATE ($\downarrow$)& mASE ($\downarrow$)& mAOE ($\downarrow$)& mAVE ($\downarrow$) & mAAE ($\downarrow$)\\
0.4637 & 0.3767 & 0.6590 & 0.2364 & 0.5139 & 0.6071 & 0.2300   \\ \hline     
\end{tabular}}

\caption{NuScenes-mini results: New model Test using the multi-dataset training.}
\label{nusc-new}
\end{table}

\begin{table}[h]
\centering
\resizebox{\columnwidth}{!}{
\begin{tabular}{ccccccc}
\hline
\multicolumn{7}{c}{Lyft Results}                                                 \\ \hline
Object Class                                  & AP & ATE & ASE & AOE & AVE & AAE \\ \hline
Car                                           & 0.389  & 0.476   & 0.217   & 0.253   & 0.703   & 0.246   \\
Bicycle                                       & 0.127  & 0.733   & 0.252   & 0.951   & 0.326   & 0.231   \\
Motorcycle                                    & 0.201  & 0.785   & 0.294   & 1.056   & 0.956   & 0.249   \\
Truck                                         & 0.246  & 0.781   & 0.198   & 0.342   & 0.468   & 0.228   \\
Bus                                           & 0.297  & 0.622   & 0.264   & 0.311   & 1.146   & 0.235   \\
Pedestrian                                    & 0.274  & 0.749   & 0.305   & 0.947   & 0.563   & 0.344   \\ \hline
\multicolumn{1}{l}{mAP for Common 6} & \multicolumn{6}{c}{0.2557}  \\ \hline
Emergency Vehicle                             & 0.108  & 0.262   & 0.437   & 1.177   & 1.098   & 0.050   \\
Other Vehicle                                 & 0.300  & 0.857   & 0.634   & 0.569   & 0.729   & 0.194   \\
Animal                                        & 0.104  & 0.304   & 0.433   & 1.089   & 0.267   & 0.399   \\ \hline
NDS ($\uparrow$) & mAP ($\uparrow$) & mATE ($\downarrow$)& mASE ($\downarrow$)& mAOE ($\downarrow$)& mAVE ($\downarrow$) & mAAE ($\downarrow$)\\
0.3677 & 0.2046 & 0.5569 & 0.2763 & 0.670 & 0.6250 & 0.2176 \\ \hline   
\end{tabular}}
\caption{Lyft results: New model Test using the multi-dataset training.}
\label{lyft-new}
\end{table}

\subsection{Ablation Studies and Analysis} 
In Table \ref{ablation} below, three experiment results are shown. And I will focus on the performance of NDS, mAP for 6 shared classes and mAP for both datasets NuScenes-mini and Lyft. 

\begin{table}[h]
\centering
\resizebox{\columnwidth}{!}{
\begin{tabular}{ccccccccc}
\hline
\multicolumn{1}{c|}{Target Domain} & \multicolumn{8}{c}{NuScenes-mini}                                                                                           \\ \hline
Method                             & NDS($\uparrow$) & mAP-6($\uparrow$) & mAP($\uparrow$) & mATE($\downarrow$) & mASE($\downarrow$) & mAOE($\downarrow$) & mAVE($\downarrow$) & mAAE($\downarrow$) \\ \hline
NuScenes-mini Only  & 0.3638  & 0.3560 &  0.3248  & 0.7729  & 0.3232   & 0.8915 & 0.6644 & 0.3332 \\
MdT        &0.3605   &0.3620    & 0.3284   & 0.8948   & 0.3203   & 0.8422  & 0.6532  & 0.3256 \\
MdT+Transformer     & 0.4637  &   0.4370  & 0.3767   & 0.6590   & 0.2364   & 0.5139   & 0.6071 & 0.2300    \\ \hline
\multicolumn{1}{c|}{Target Domain} & \multicolumn{8}{c}{Lyft}                                                                                                     \\ \hline
Method                             & NDS($\uparrow$) & mAP-6($\uparrow$)  & mAP($\uparrow$) & mATE($\downarrow$) & mASE($\downarrow$) & mAOE($\downarrow$) & mAVE($\downarrow$) & mAAE($\downarrow$) \\ \hline
NuScenes-mini Only     &  0.0012  &0.004   & 0.0024   &1.0  &  1.0   & 1.0  &1.0   &  1.0        \\
MdT       &   0.2541     & 0.1658     & 0.1044    & 0.5460    & 0.3647   & 0.4496  & 0.3914 & 0.2670                \\
MdT+Transformer    &  0.3677        & 0.2557    & 0.2046     & 0.5569  & 0.2763   & 0.670    & 0.6250  & 0.2176 \\ \hline              
\end{tabular}}

\caption{Ablation Study on Multi-dataset Training (MdT here) and the Transformer. There are three experiment results in total: Base model trained on NuScenes-mini,  Base model using Multi-dataset training for dataset transfer and the new efficient network with MdT. The mAP-6 represents the 6 shared classes between NuScenes-mini and Lyft.}
\label{ablation}
\end{table}

\subsubsection{Multi-source Domain Adaptation}
In Tabel \ref{ablation}, we can see that for the two test results of the NuScenes-mini part, the changes in NuScenes-mini training and multi-dataset training are very weak. This is because my multi-dataset itself is Contains all NuScenes-mini. This small change is not surprising, because this part is mainly used for the transfer capability of training models, facing data and domains like Lyft that are several times larger than NuScenes-mini. Looking at it this way, the improvement brought by multi-dataset training technology is considerable. The test data on Lyft shows that NDS increased by 0.25 (0.0012$\rightarrow$0.2541), and mAP-6 increased by 0.16 (0.04$\rightarrow$0.1658). mAP increased by 0.10 (0.024$\rightarrow$0.1044). The main increase comes from mAP-6.

\subsubsection{Transformer for 3D Object Detection Tasks}
For the Transformer structure I designed, the performance improvement it provides is very significant, even NuScenes-mini's NDS (0.3605$\rightarrow$0.4637), mAP-6 (0.3620$\rightarrow$0.4370) and mAP (0.3284$\ rightarrow $0.3767) exceeds the impact of multi-dataste training. Lyft's improvement here is also amazing. However, there is no way to further verify whether the changes in Lyft's NDS (0.2541$\rightarrow$0.3677), mAP-6 (0.2658$\rightarrow$0.2577) and mAP (0.1044$\rightarrow$0.2046) come from changes in the network structure or the training data Increase and diversify. 

\subsubsection{Network Effectiveness}
My lightweight model has demonstrated high efficiency, and I think it has the potential to become a cornerstone choice for handling 3D object detection tasks in the future. The computational efficiency of the model is a key feature. By adopting the network architecture of multi-layer feature and Transformer models and efficient algorithms, my model can be trained and inferred with only one GTX2080Ti (12GB). This not only reduces hardware requirements but also shortens training time, which will not exceed 8 hours at a time. This facilitates rapid iteration and model improvement. Second, we have made significant progress in resource efficiency. The entire training and testing requires no more than 100GB of memory, which means it can run in constrained hardware environments and does not put unnecessary pressure on system resources.

\section{Conclusion}
\label{sec:conclusion}
 I proposed a network structure for multi-view 3D object detection. Through multi-dataset training and anchor frame detection head based on the Transformer structure, the network demonstrated good domain transfer performance and efficient detection performance by image features and 3D information combined. The purpose of the network is to use a small amount of data in a limited source domain and use the existing large model pre-training backbone weights to achieve competitive index data in the new target domain. My main techniques include a diverse 3D dataset that fuses different datasets across domains, 2D multi-level image feature fusion, new anchor box object query, utilizing 3D information as available semantic information, and 2D multi-view image feature blending with 3D free anchor box. In conclusion, my network achieved good results on standard metrics of 3D object detection by using data transfer, and it shows promising potential in '2D features-3D space' for 3D visual recognition by using 2D visual recognition techniques.

\bibliographystyle{ieeetr}
\bibliography{11_references}

\ifarxiv \clearpage \appendix \section{Appendix Section}
Supplementary material goes here.
 \fi

\end{document}


\title{\paperTitle}
\author{\authorBlock}
\maketitlesupplementary

\section{Appendix Section}
Supplementary material goes here.

{\small
\bibliographystyle{ieee_fullname}
\bibliography{11_references}
}